\documentclass{article}

\usepackage{arxiv}

\usepackage{graphicx}
\usepackage{amsmath}
\usepackage{amssymb}
\usepackage{booktabs}
\usepackage{bbm}
\usepackage{url}
\usepackage{multirow,makecell, rotating}
\usepackage{booktabs}
\usepackage{tikz}
\usepackage{caption}
\usepackage{graphbox}
\usepackage{comment}
\usepackage{epsfig}
\usepackage{bm}
\usepackage{soul}
\usepackage{color}
\usepackage{wrapfig}
\newtheorem{definition}{Definition}
\usepackage{subcaption}

\usepackage[pagebackref=true,breaklinks=true,letterpaper=true,colorlinks,bookmarks=false]{hyperref} 

\usepackage[capitalize]{cleveref}
\crefname{section}{Sec.}{Secs.}
\Crefname{section}{Section}{Sections}
\Crefname{table}{Table}{Tables}
\crefname{table}{Tab.}{Tabs.}

\begin{document}

\title{Con\textit{ffusion}: Confidence Intervals for Diffusion Models}

\author{Eliahu Horwitz, Yedid Hoshen\\
  School of Computer Science and Engineering\\
  The Hebrew University of Jerusalem, Israel\\
  \url{https://www.vision.huji.ac.il/conffusion}\\
  \texttt{\{eliahu.horwitz, yedid.hoshen\}@mail.huji.ac.il} \\
}

\maketitle

\begin{figure*}[h]
    \begin{tabular}{@{\hskip1pt}c@{\hskip1pt}c@{\hskip1pt}c@{\hskip1pt}c@{\hskip1pt}c@{\hskip1pt}c}
        \includegraphics[width=0.16\linewidth]{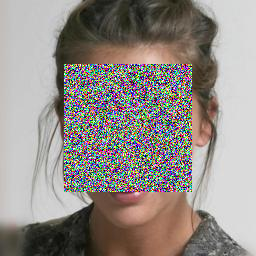} & 
        \includegraphics[width=0.16\linewidth]{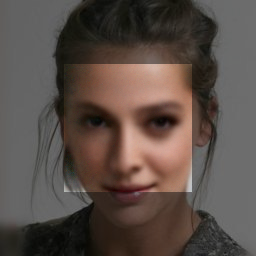} &
        \includegraphics[width=0.16\linewidth]{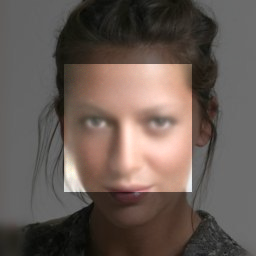} & 
        \includegraphics[width=0.16\linewidth]{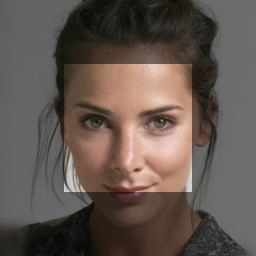} & 
        \includegraphics[width=0.16\linewidth]{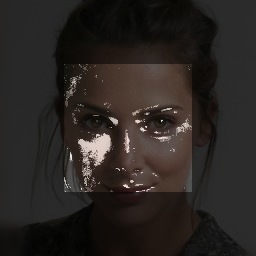} & 
        \includegraphics[width=0.16\linewidth]{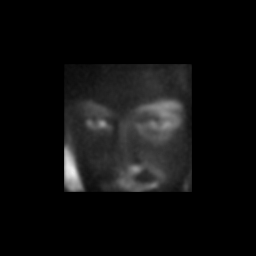} \\ 

        \includegraphics[width=0.16\linewidth]{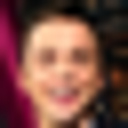} & 
        \includegraphics[width=0.16\linewidth]{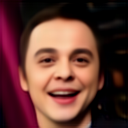} &
        \includegraphics[width=0.16\linewidth]{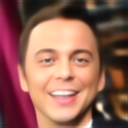} & 
        \includegraphics[width=0.16\linewidth]{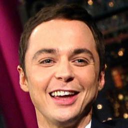} & 
        \includegraphics[width=0.16\linewidth]{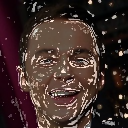} & 
        \includegraphics[width=0.16\linewidth]{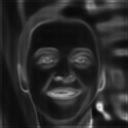}  \\
        
        Input & Lower Bound & Upper Bound & Ground Truth & Error & Interval Sizes \\
    \end{tabular}
 \caption{\textit{\textbf{Con\textit{ffusion}:}} Given a corrupted input image, our method ``Con\textit{ffusion}``, repurposes a pretrained diffusion model to generate lower and upper bounds around each reconstructed pixel. The true pixel value is guaranteed to fall within these bounds with probability $p$. We present the bounds for inpainting (top row, the context is dimmed for visualization) and super-resolution (bottom row). Tighter intervals provide more information; we visualize the normalized interval size in the rightmost column, darker values are tighter intervals}
\label{fig:rgb_v_d}
\end{figure*}

\begin{abstract}
Diffusion models have become the go-to method for many generative tasks, particularly for image-to-image generation tasks such as super-resolution and inpainting. Current diffusion-based methods do not provide statistical guarantees regarding the generated results, often preventing their use in high-stakes situations. To bridge this gap, we construct a confidence interval around each generated pixel such that the true value of the pixel is guaranteed to fall within the interval with a probability set by the user. Since diffusion models parametrize the data distribution, a straightforward way of constructing such intervals is by drawing multiple samples and calculating their bounds. However, this method has several drawbacks: i) slow sampling speeds ii) suboptimal bounds iii) requires training a diffusion model per task. To mitigate these shortcomings we propose Con\textit{ffusion}, wherein we fine-tune a pre-trained diffusion model to predict interval bounds in a single forward pass. We show that Con\textit{ffusion} outperforms the baseline method while being three orders of magnitude faster.

\end{abstract}

\section{Introduction}
Diffusion models (DMs) have become the primary method for a wide range of generative tasks such as super-resolution\cite{sr3}, inpainting, and colorization \cite{palette}. However, as is the case with many other deep-learning-based methods, they are still somewhat of a black box. Deploying DMs in real-world, high-stakes situations requires a way of statistically guaranteeing the degree of confidence\footnote{The terms ``confidence`` and ``uncertainty`` are often used interchangeably, in this paper, we opt to use the term confidence.} the model has in its prediction. To this end, we construct an interval around each generated pixel such that the true value of the pixel lies within the interval with a probability set by the user. As an example, a doctor analysing a low-resolution MRI scan may greatly benefit from increasing its resolution. However, as generative models tend to ``hallucinate`` details, the doctor might not be able to trust that a high-resolution image generated by the latest diffusion model is, in fact, true to reality. Providing the doctor with a statistically guaranteed interval around each pixel will allow the doctor to use the output of the generative model. 

In this paper we use diffusion models for constructing confidence intervals in image-to-image tasks (e.g. super-resolution and inpainting). A naive approach can directly use the excellent distribution approximation capabilities of diffusion models. It would first sample from the model multiple solutions given an input image. It will use the distribution of solutions to estimate the bounds on each pixel. Despite the simplicity of this method, we find that it yields intervals that are tighter than Angelopoulos et al. \cite{im2imuq}, the current leading method for interval construction that makes use of quantile regression\cite{quantile_regression}.
The naive sampling method is very slow, as multiple sampled variations are needed for each test image, and given that a single DM inference step requires multiple denoising steps. In fact, it requires thousands of forward passes for each test image. To accelerate the sampling bottleneck we introduce an amortized method that approximates the sampled bounds in a single forward step and without sampling multiple variations at inference time. Unfortunately, we find that approximating the sampled bounds often yields suboptimal intervals.

We propose Con\textit{ffusion}, a method for combining the best of both worlds. Given a pretrained DM, we fine-tune it to the task of bound estimation using the quantile regression loss. Although diffusion models usually require multiple forward passes for a single generation, Con\textit{ffusions} only require a single forward pass. We achieve this by decoupling the denoising model from the diffusion process. Moreover, owing to the strong pretrained features, we demonstrate that a generic DM (e.g. ADM \cite{ADM}) pretrained on ImageNet\cite{imagenet} can be finetuned to extract confidence intervals on different datasets (e.g. CelebA-HQ\cite{celeba_hq}) and different tasks (e.g. super-resolution and inpainting).

To summarize, our main contributions are:
\begin{enumerate}
    \item Equipping diffusion models with pixel-wise confidence intervals, thus providing rigorous statistical guarantees on the prediction.
    \item Proposing Con\textit{ffusion}, a method combining diffusion pretraining with quantile regression for confidence interval construction in a single forward pass.
    \item Empirically demonstrating the efficacy of diffusion pretraining for confidence interval construction. Our method achieves tighter intervals compared to the current state-of-the-art while providing a speed-up of three orders of magnitude over the sampling baseline.
\end{enumerate}

\begin{figure*}[t]
\begin{tabular}{@{\hskip1pt}c@{\hskip1pt}c@{\hskip1pt}c@{\hskip1pt}c@{\hskip1pt}c@{\hskip1pt}c}

\includegraphics[width=0.16\linewidth]{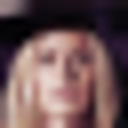} & 
\includegraphics[width=0.16\linewidth]{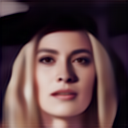} & 
\includegraphics[width=0.16\linewidth]{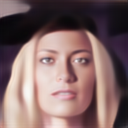} &
\includegraphics[width=0.16\linewidth]{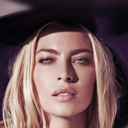} & 
\includegraphics[width=0.16\linewidth]{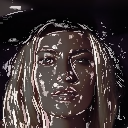} &
\includegraphics[width=0.16\linewidth]{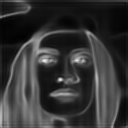} \\
\includegraphics[width=0.16\linewidth]{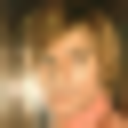} & 
\includegraphics[width=0.16\linewidth]{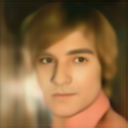} & 
\includegraphics[width=0.16\linewidth]{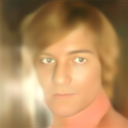} & 
\includegraphics[width=0.16\linewidth]{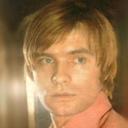} &
\includegraphics[width=0.16\linewidth]{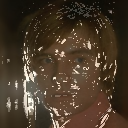} &
\includegraphics[width=0.16\linewidth]{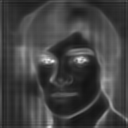}\\
\includegraphics[width=0.16\linewidth]{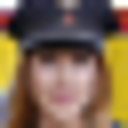} & 
\includegraphics[width=0.16\linewidth]{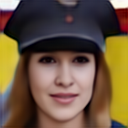} & 
\includegraphics[width=0.16\linewidth]{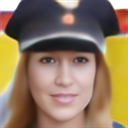} & 
\includegraphics[width=0.16\linewidth]{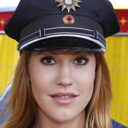} &
\includegraphics[width=0.16\linewidth]{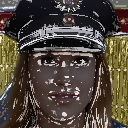} & 
\includegraphics[width=0.16\linewidth]{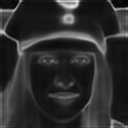}\\
Input & Lower Bound & Upper Bound & Ground Truth & Error & Interval Size \\
\end{tabular}
 \caption{\textit{\textbf{Super-resolution:}} We present the bounds for super-resolution extracted via N-Con\textit{ffusion}. When needed, the bounds span a wide range, e.g., bounding eye color from below with darker colors and from above with brighter ones. As expected, areas with higher frequencies contain more errors and have wider intervals (e.g. hair)}
\label{fig:sr_fig}
\end{figure*}

\section{Related Work}
\subsection{Diffusion Models}
Originating from the field of score-based models \cite{score_matching}, recently, Denoising Diffusion Probabilistic Models (DDPMs)\cite{DDPM} have emerged as a new paradigm for image generation, surpassing GANs\cite{GAN} and achieving state-of-the-art results\cite{ADM}. Using a U-Net\cite{unet} to gradually denoise the samples, they learn the distribution of the training set. DDIM \cite{DDIM} extends the stochastic sampling of DDPMs to allow for deterministic sampling. The recent EDM \cite{EDM} reformulates the theory behind diffusion models and decouples it from design choices. DMs have been used for many applications such as image editing \cite{blended_diffusion, diffae, blended_latent_diffusion} and text-to-image generation \cite{imagen, dalle2, ldm, prompt2prompt, dreambooth, text_inv}. Most related to our work are applications in inverse problems such as super-resolution and inpainting \cite{palette, repaint, cascaded, come_closer, srdiff, sr3}.

\subsection{Confidence Guarantees}
Deriving statistically rigorous confidence intervals/sets for predictions of machine learning models is an established field of research \cite{conformal_feat, conformal_noise, limits_conformal, bates2021distribution}. One popular paradigm for quantifying the confidence of a model with a user-defined probability guarantee is ``Conformal Prediction``, an introduction to the field is available in \cite{intro2conformal}. Our work is most closely related to \cite{im2imuq}, in which the authors extend the conformal prediction framework to image-to-image regression problems. Differently than \cite{im2imuq}, our method leverages a pretrained diffusion model for extracting tighter bounds at a single forward pass. The goal is to construct an interval around each generated pixel such that the true value is guaranteed to lie within the interval with a probability specified by the user. Quantile regression \cite{quantile_regression} is used to train the model to output upper and lower bounds, after training, a calibration phase adjusts the bounds so that they are as tight as possible while adhering to the restriction set by the user. \cite{disentabgled_semantic_intervals} use a similar setting for deriving semantic intervals.

\section{Preliminaries}
\label{sec:preliminaries}
\subsection{Diffusion Models}
Diffusion models convert sampled Gaussian noise $\epsilon$ into a sample $x_0$ from the empirical data distribution. This mapping is performed by learning a neural denoising network $f_{\theta}$ to gradually reverse the noising process (aka diffusion) from $x_0$ to $x_t$. Conditional diffusion models \cite{sr3, palette, ldm} condition the denoising process on an input signal. Specifically, image-to-image DMs perform the $p(y|x)$ denoising step where both $x$ and $y$ are images (e.g. $y$ is the target high-resolution image and $x$ is the source low-resolution image). The DM generates a target image $y_0$ over $T$ denoising steps. Starting from isotropic Gaussian noise $y_T \sim \mathcal{N}(0,I)$, the model iteratively refines the image using the learned transition distribution $p_{\theta}(y_{t-1}|y_t,x)$ until arriving at $y_0 \sim p(y|x)$. This iterative process is time and compute-intensive as image generation requires multiple network evaluations. In Sec.~\ref{sec:method} we show how Con\textit{ffusions} avoid this with a relaxed prediction objective.
Most image DMs use a U-Net to learn the denoising step, the current step $t$ is encoded and passed into the network alongside $y_t$. When conditioning on images, instead of starting from a Gaussian noise image, we often diffuse the input image only to some $t\in[0,T)$, this is in order to preserve some of the original input signal.

\subsection{Confidence Guarantees}
\label{sec:conf_pre}
\noindent \textbf{Definitions.} We denote a calibration set $\{x_i,y_i\}_{i=1}^{N}$ where $x_i, y_i~\in~[0,1]^{M\times N}$ are a corrupted and target image respectively. Our goal is to construct a confidence interval around each pixel of $\hat{y_i}$ such that the true value of the pixel lies within the interval with a probability set by the user. Formally, for each pixel we construct the following interval 

\begin{equation}
\label{eq:interval}
    \mathcal{T}(x_{i_{mn}}) = \left[\hat{l}(x_{i_{mn}}), \hat{u}(x_{i_{mn}})\right]
\end{equation}

where $\hat{l}, \hat{u}$ are the interval lower and upper bounds. To provide the interval with statistical soundness, the user selects a risk level $\alpha\in(0,1)$ and an error level $\delta\in(0,1)$. We then construct intervals such that at least $1-\alpha$ of the ground truth pixel values are contained in it with probability $1-\delta$. That is, with probability of at least $1-\delta$,

\begin{equation}
\label{eq:guarantee}
    \mathbb{E}\left[\frac{1}{MN}\Big|\big\{(m,n) : y_{(m,n)} \in \mathcal{T}(x)_{(m,n)}\big\}\Big|\right] \ge 1 - \alpha,
\end{equation}

where $x, y$ are a test sample and label originating from the same distribution as the calibration set. In Sec.~\ref{sec:method} we introduce four different methods for constructing these intervals with $\tilde{l}$ and $\tilde{u}$ serving as approximations for $\hat{l}$ and $\hat{u}$.\\

\noindent As we have made no explicit assumptions about $\tilde{l}$ and $\tilde{u}$, they are merely heuristics and do not provide an actual statistical guarantee. To this end, we scale (aka calibrate) $\tilde{l}$ and $\tilde{u}$ by a constant $\hat{\lambda}$ such that they satisfy \ref{eq:guarantee}, adhering to the risk and error levels set by the user. We denote the set of all intervals for each pixel of each image $[\hat{\lambda}\tilde{L}, \hat{\lambda}\tilde{U}]=[\hat{L}, \hat{U}]$. Following \cite{bates2021distribution, im2imuq}, we call this set an image-valued \textit{Risk-Controlling Prediction Set}.

\begin{definition}[Risk-Controlling Prediction Set (RCPS) \cite{bates2021distribution, im2imuq}]
\label{def:rcps}
We call a random set-valued function $\mathcal{T}:\mathcal{X}\to\left(2^{[0,1]}\right)^{M \times N}$
    an \emph{($\alpha$,$\delta$)-Risk-Controlling Prediction Set} if 
    \begin{equation}
        \label{eq:rcps}
        \mathbb{P}\left(\mathbb{E}\big[L(\mathcal{T}(x),y)\big] > \alpha \right) \leq \delta,
    \end{equation}
    where 
    \begin{equation}
        L(\mathcal{T}(X),Y) = 1-\frac{\Big|\big\{(m,n) : y_{(m,n)} \in \mathcal{T}(x)_{(m,n)}\big\}\Big|}{MN}.
    \end{equation}

\end{definition}
Where the inner expectation in \ref{eq:rcps} is over a new test image $(x,y)$ and the outer probability is over the calibration set  $\{x_i,y_i\}_{i=1}^{N}$. Meaning, the RCPS from Def.~\ref{def:rcps} only fails to control the risk with probability $\delta$.

\noindent \textbf{Calibration.}
Thus far we have made no assumptions regarding $\tilde{l}$ and $\tilde{u}$, as mentioned above, this guarantee is provided by the calibration process. During calibration, we pick the smallest intervals such that the number of pixels falling outside is below $\alpha$. We are interested in the smallest $\lambda$ such that $\mathcal{T}_{\lambda} =[\lambda\tilde{l}, \lambda\tilde{u}]$ satisfies our RCPS definition. Hence, starting from a large enough $\lambda$ we iteratively construct $\mathcal{T}_{\lambda}$ until Def.~\ref{def:rcps} is no longer met, resulting in the calibration constant $\hat{\lambda}$. In this work we use the \textit{validation} set for computing $\hat{\lambda}$ and report the results on intervals extracted from the \textit{test} set, scaled by $\hat{\lambda}$. Further details about the specific calibration algorithm used are available in App.~\ref{app:calibration} as well as in \cite{im2imuq, bates2021distribution}.

\begin{figure*}[t]
\begin{tabular}{@{\hskip1pt}c@{\hskip1pt}c@{\hskip1pt}c@{\hskip1pt}c@{\hskip1pt}c@{\hskip1pt}c}

\includegraphics[width=0.16\linewidth]{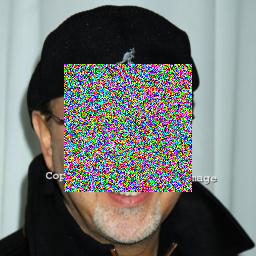} & 
\includegraphics[width=0.16\linewidth]{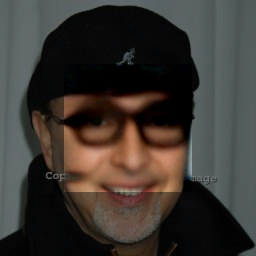} & 
\includegraphics[width=0.16\linewidth]{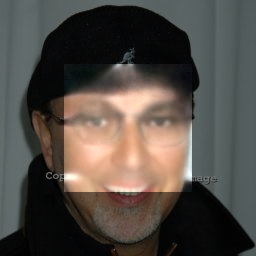} &
\includegraphics[width=0.16\linewidth]{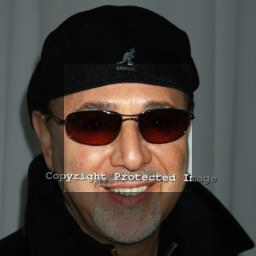} & 
\includegraphics[width=0.16\linewidth]{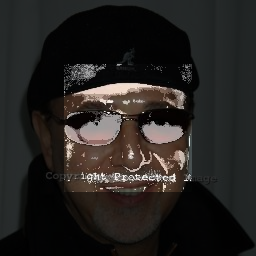} &
\includegraphics[width=0.16\linewidth]{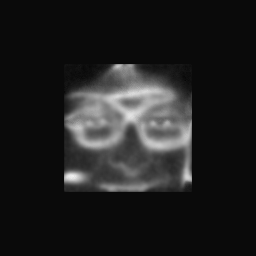} \\
\includegraphics[width=0.16\linewidth]{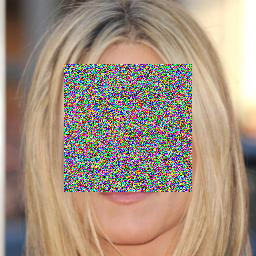} & 
\includegraphics[width=0.16\linewidth]{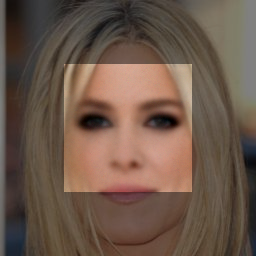} & 
\includegraphics[width=0.16\linewidth]{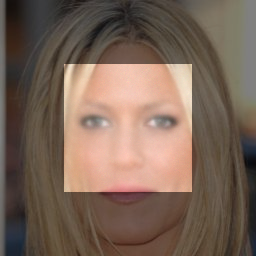} & 
\includegraphics[width=0.16\linewidth]{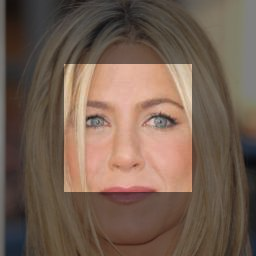} &
\includegraphics[width=0.16\linewidth]{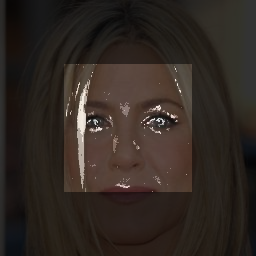} &
\includegraphics[width=0.16\linewidth]{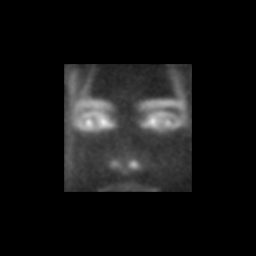}\\
\includegraphics[width=0.16\linewidth]{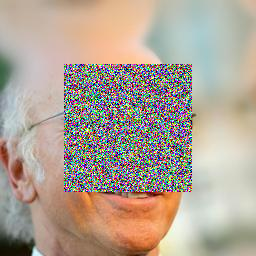} & 
\includegraphics[width=0.16\linewidth]{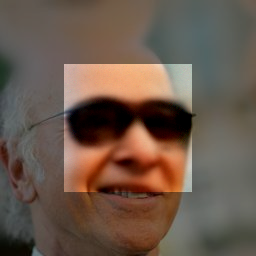} & 
\includegraphics[width=0.16\linewidth]{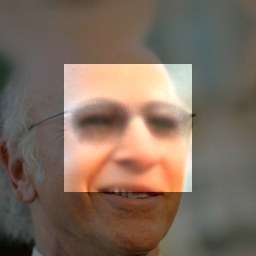} & 
\includegraphics[width=0.16\linewidth]{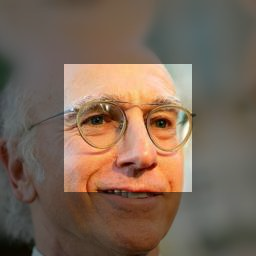} &
\includegraphics[width=0.16\linewidth]{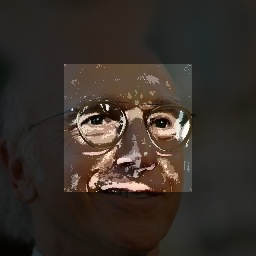} & 
\includegraphics[width=0.16\linewidth]{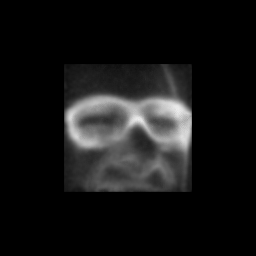}\\
Input & Lower Bound & Upper Bound & Ground Truth & Error & Interval Size \\
\end{tabular}
 \caption{\textit{\textbf{Inpainting:}} We present the bounds for inpainting (context is dimmed for visualization) extracted via N-Con\textit{ffusion}. In areas with a wider distribution, our bounds cover the distribution. Bounding from below with sunglasses and darker eyes and from above with eyeglasses and brighter eyes. The interval size reveals areas where the model is less confident in its prediction (e.g. glasses and eyes)}
\label{fig:inpainting_fig}
\end{figure*}

\section{Method}
\label{sec:method}
We now describe different methods for constructing confidence intervals.
\subsection{A Baseline for Interval Construction} 
The recent ``im2im-uq``\cite{im2imuq} proposed a unified interval construction schema for regression tasks. They train a neural network to predict three outputs for a given input image, the upper and lower bounds and the pointwise reconstruction. To estimate the bounds they use the well-known \textit{quantile loss} (sometime also called the \textit{pinball loss}) shown below,
    \begin{equation}
        \mathcal{L}_{\alpha}\big(\hat{q}_{\alpha}(x),y\big) = \big(y-\hat{q}_{\alpha}(x)\big)\alpha\mathbbm{1}\{y > \hat{q}_{\alpha}(x)\} 
        + \big(\hat{q}_{\alpha}(x) - y\big)(1-\alpha)\mathbbm{1}\{y \leq \hat{q}_{\alpha}(x)\}
    \end{equation}
    
    where $\alpha$ is the quantile to optimize and $\hat{q}_{\alpha}(x)$ its quantile estimator. Since we require $\tilde{u}$ and $\tilde{l}$ to estimate different quantiles, the quantile loss becomes,
\begin{equation}
\label{eq:qr_loss}
    \mathcal{L}_{QR}(x,y)=\mathcal{L}_{\alpha/2}\big(\tilde{l}(x),y\big) + \mathcal{L}_{1-\alpha/2}\big(\tilde{u}(x),y\big)
\end{equation}

the final objective combines the quantile loss for bound estimation with \textit{MSE} for pointwise prediction, as shown below,

\begin{equation}
\label{eq:im2imuq}
    \mathcal{L}(x,y)= \mathcal{L}_{QR}(x,y) + \mathcal{L}_{mse}(x,y)
\end{equation}

After training, $\tilde{u}$ and $\tilde{l}$ should approximate $1-\alpha/2$ and $\alpha/2$, the calibration discussed in Sec.\ref{sec:conf_pre} is used to obtain the final bounds $\hat{u}$ and $\hat{l}$. 
Following \cite{im2imuq} this baseline method does not use pretraining, this will be further elaborated on in the following sections. We use this method as our baseline and refer to it as $ADM_{UQ}$.

\subsection{Sampling-based Bounds Estimation}
\noindent \textbf{Sampling-based Bounds.}
Owing to their excellent distribution approximation capabilities, DMs are natural candidates for bound estimation. Thus, we can construct intervals by sampling from the approximated distribution of a pretrained DM. Specifically, given a DM trained on a specific restoration task (e.g. super-resolution, inpainting), we sample multiple reconstructed variations $SV_i=\{\hat{y_{i_{j}}}\}_{j=1}^{J}$ for each corrupted calibration item $x_i$, thus approximating the image distribution. We extract the interval bounds by taking an upper and lower quantile of each pixel in $SV_i$. For example, assuming $\alpha=0.1$ (i.e. we allow for at most $10\%$ errors) we have $\tilde{l}=\alpha/2=0.05$ and $\tilde{u}=1-\alpha/2=0.95$ as the initial interval bounds. Using the set of bounds $\tilde{U}, \tilde{L}$ extracted from all the images, we perform the calibration described in Sec.~\ref{sec:conf_pre} to select $\hat{\lambda}$ such that $\hat{\lambda}\tilde{U}=\hat{U}$ and $\hat{\lambda} \tilde{U} = \hat{U}$. At inference time, given a new corrupted input $x$, we construct a \textit{Sampled Variation} set $SV$ and extract $\tilde{u}, \tilde{l}$. Finally, we use the $\hat{\lambda}$ from the calibration phase and return $\hat{\lambda} \tilde{u}=\hat{u}$ and $\hat{\lambda} \tilde{l}=\hat{l}$. 
This method yields tighter bounds than those constructed by $ADM_{UQ}$, we refer to it as $DM_{SB}$. However, although estimating the bounds via sampling outperforms $ADM_{UQ}$, it requires us to sample multiple variations of each item, both during calibration and during inference. Given the need for multiple denoising steps to generate each variation, extracting bounds for a single inference image takes thousands of forward passes, rendering it infeasible for real-world use. 

\noindent \textbf{Accelerating Sampled Bounds.} Following the success of $DM_{SB}$ in generating tight intervals, we propose a simple acceleration method. Similarly to $DM_{SB}$, given a DM trained for a specific restoration task, we construct $SV_i$ and extract the set of bounds denoted by $\tilde{U}_{sv}, \tilde{L}_{sv}$. To speed up inference, we finetune the original DM such that given an input image $x_i$, we predict $\tilde{u}_i$ and $\tilde{l_i}$ in a single forward step. Specifically, denote the pretrained DM with $\mathcal{DM}$, our finetuning objective is,

\begin{equation}
    \mathcal{L}\big(x,\tilde{l}_{sv},\tilde{u}_{sv}\big)=\mathcal{L}_{mse}\big(\mathcal{DM}(x),\tilde{l}_{sv}\big) + \mathcal{L}_{mse}\big(\mathcal{DM}(x),\tilde{u}_{sv}\big)
\end{equation}

It is somewhat surprising that we are able to approximate the sampled bounds in a single forward pass, seemingly bypassing the original motivation of the diffusion training objective. However, we conjecture this success is owed to the relaxation in the training objective. The diffusion process allows for a high level of fidelity and detail and excels in generating high frequencies, as interval bounds contain mostly lower frequencies, they are easier to generate in a single step. Note that differently from $DM_{SB}$, here we construct a set of bounds both for the \textit{finetune} and \textit{validation} sets. We denote this method as $DM_{SBA}$.

\subsection{Con\textit{ffusions}}
Although accelerating the sampled bounds produces reasonable results, it still requires many forward steps for constructing the calibration set. Moreover, it may not reach the same performance as the original sampling method. 
To bridge this gap, we now describe our method, Con\textit{ffusion} for adapting quantile regression (QR) to diffusion models, taking the best of both worlds.\\

\noindent \textbf{Narrow Con\textit{ffunsion} (N-Con\textit{ffusion}).} We replace the approximation of the sampled bounds used in $DM_{SBA}$ with QR. In particular, given a DM trained for a specific restoration task, we finetune the model with the $\mathcal{L}_{QR}$ objective described in Eq.~\ref{eq:qr_loss}. This combines the powerful distributional understanding of the pretrained denoising model, with the simple objective of QR. It also decouples the denoising model from the computationally demanding diffusion process. 
N-Con\textit{ffusion}, outperforms  $DM_{SB}$ in terms of interval size while still maintaining the speed of  $DM_{SBA}$, allowing for a single-step interval construction.

\noindent \textbf{Global Con\textit{ffunsion} (G-Con\textit{ffusion}).} While N-Con\textit{ffusion} outperforms all previously discussed methods, it requires a DM pertained to a specific data modality for the specific task. This is a limitation, as DMs are difficult to train and require large amounts of compute and data. We thus propose G-Con\textit{ffusion}, a simple extension that makes Con\textit{ffusions} data and task agnostic. Instead of using task-specific models such as SR3\cite{sr3} and Palette \cite{palette}, we use a large-scale DM pretrained for general image generation (unconditional or class conditional). In particular, we use ADM \cite{ADM} (Guided Diffusion) pretrained on ImageNet. We can perform multiple tasks given a single pretrained model by finetuning ADM with the same QR objective from Eq.~\ref{eq:qr_loss}. This saves the cost of developing specialized models for new tasks and data modalities.

\begin{figure*}[t]
\begin{tabular}{c@{\hskip1pt}c@{\hskip1pt}c@{\hskip1pt}c@{\hskip1pt}c@{\hskip1pt}c@{\hskip1pt}c@{\hskip1pt}}

\rotatebox[ origin=c]{90}{$ADM_{UQ}$} &
\includegraphics[align=c, width=0.16\linewidth]{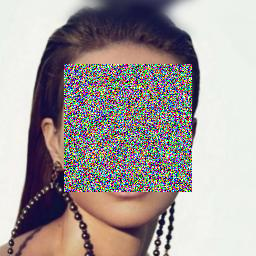} & 
\includegraphics[align=c, width=0.16\linewidth]{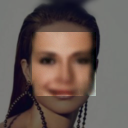} & 
\includegraphics[align=c, width=0.16\linewidth]{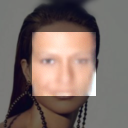} & 
\includegraphics[align=c, width=0.16\linewidth]{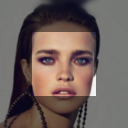} & 
\includegraphics[align=c, width=0.16\linewidth]{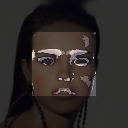} &
\includegraphics[align=c, width=0.16\linewidth]{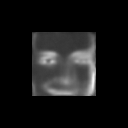}\\\addlinespace

\rotatebox[ origin=c]{90}{$DM_{SB}$} &
\includegraphics[align=c, width=0.16\linewidth]{figs/compare/SB_406_masked_smpl.png} & 
\includegraphics[align=c, width=0.16\linewidth]{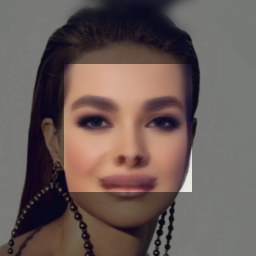} & 
\includegraphics[align=c, width=0.16\linewidth]{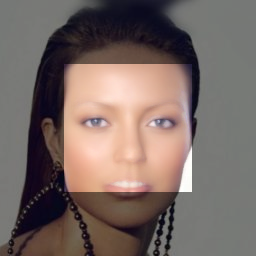} & 
\includegraphics[align=c, width=0.16\linewidth]{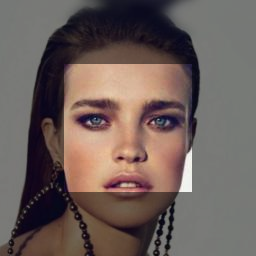} & 
\includegraphics[align=c, width=0.16\linewidth]{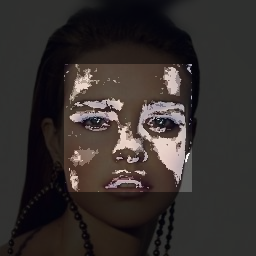} &
\includegraphics[align=c, width=0.16\linewidth]{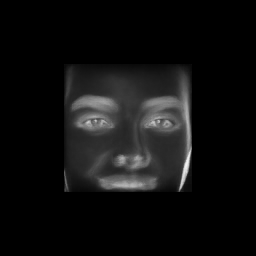}\\\addlinespace

\rotatebox[ origin=c]{90}{$DM_{SBA}$} &
\includegraphics[align=c, width=0.16\linewidth]{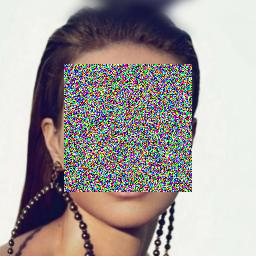} & 
\includegraphics[align=c, width=0.16\linewidth]{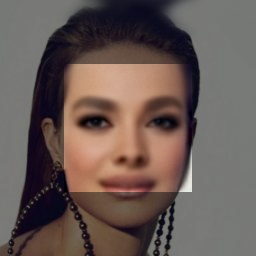} & 
\includegraphics[align=c, width=0.16\linewidth]{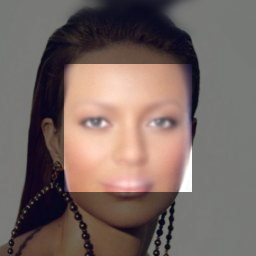} & 
\includegraphics[align=c, width=0.16\linewidth]{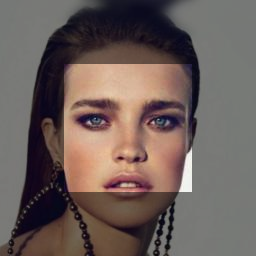} & 
\includegraphics[align=c, width=0.16\linewidth]{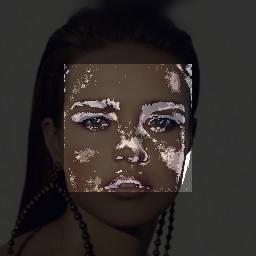}&
\includegraphics[align=c, width=0.16\linewidth]{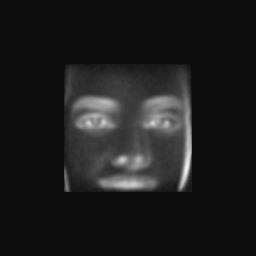}\\\addlinespace

\rotatebox[ origin=c]{90}{N-Con\textit{ffusion}} &
\includegraphics[align=c, width=0.16\linewidth]{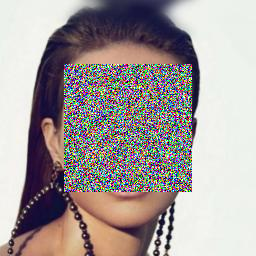} & 
\includegraphics[align=c, width=0.16\linewidth]{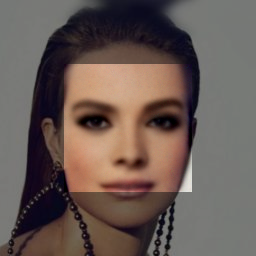} & 
\includegraphics[align=c, width=0.16\linewidth]{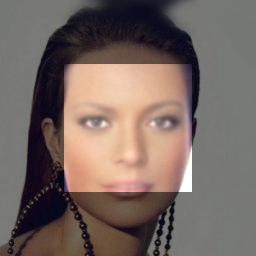} & 
\includegraphics[align=c, width=0.16\linewidth]{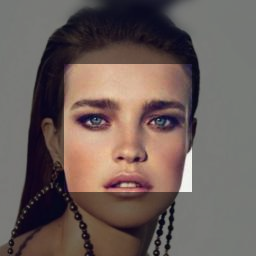} & 
\includegraphics[align=c, width=0.16\linewidth]{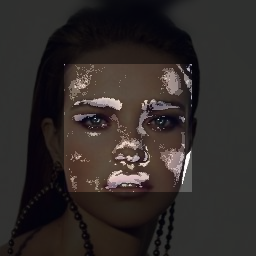} &
\includegraphics[align=c, width=0.16\linewidth]{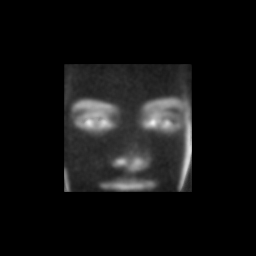}\\\addlinespace

\rotatebox[ origin=c]{90}{G-Con\textit{ffusion}} &
\includegraphics[align=c, width=0.16\linewidth]{figs/compare/SB_406_masked_smpl.png} & 
\includegraphics[align=c, width=0.16\linewidth]{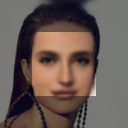} & 
\includegraphics[align=c, width=0.16\linewidth]{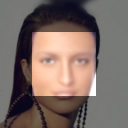} & 
\includegraphics[align=c, width=0.16\linewidth]{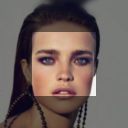} & 
\includegraphics[align=c, width=0.16\linewidth]{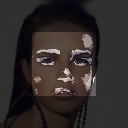} &
\includegraphics[align=c, width=0.16\linewidth]{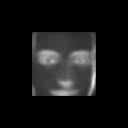}\\\addlinespace
 & Input & Lower Bound & Upper Bound & Ground Truth & Error & Interval Size \\
\end{tabular}
 \caption{\textit{\textbf{Comparing the Different Methods:}} We compare the different methods on the inpainting task. $ADM_{UQ}$ produces blurry bounds, also apparent from the smoother interval heatmap. Although $DM_{SB}$ generates the sharpest intervals, the estimated bounds may contain artifacts resulting from the lack of interpolation capabilities (e.g. ``duplicated``  nose and lips in the lower bound). N-Con\textit{ffusion} combines the best of both worlds, generating sharp intervals while maintaining realistic bounds}
\label{fig:compare}
\end{figure*}

\section{Experiments}
\label{sec:experiments}
\subsection{Evaluation Metrics}
\noindent \textbf{Empirical Risk.}
The first metric is the empirical risk (i.e. error), the risk is calculated as the percentage of pixel values that fall outside of their predicted interval. Following \ref{eq:guarantee} and Def.~\ref{def:rcps}, the risk should fall below $\alpha$ with probability $1-\delta$.

\noindent \textbf{Interval Size.}
A trivial solution that satisfies the statistical guarantee is the interval $[0,1]$. However, such an interval is useless, tighter intervals provide more insight into the quality of the prediction. Hence, we report the mean interval size, striving for tighter intervals.

\noindent \textbf{Size-stratified Risk.}
In addition to tight bounds, we desire intervals whose errors are uniformly distributed across pixels. For example, a model with non-uniform errors could predict a fixed interval for $1-\alpha$ pixels, in which case the intervals are no longer useful in measuring confidence. To this end we report the size-stratified risk \cite{intro2conformal}; we divide the pixels into quartiles of interval sizes and report the empirical risk for each quartile. We strive for quartiles that have approximately the same risk. This is visualized by a bar plot, where the closer the height of the bars, the better the result is, meaning the model does not include easier areas to make up for challenging ones.

\noindent \textbf{Visualizations.} In addition to quantitative evaluations, we visualize the resulting intervals. We provide three types of visualizations: i) the lower and upper bounds of each pixel in the image ii) a risk heatmap, visualizing intervals that do not contain the true value iii) a heatmap of the interval lengths, for visualization purposes the values are normalized. When visualizing inpainting, we dim the area that is not inpainted so that the context and inpainted areas are well separated.

\subsection{Implementation Details}
\noindent \textbf{Models.} We evaluate the constructed intervals on the inpainting and super-resolution tasks. For super-resolution we use the publicly available implementation\footnote{\url{https://github.com/Janspiry/Image-Super-Resolution-via-Iterative-Refinement}} of SR3\cite{sr3} pretrained on FFHQ\cite{ffhq} to perform $16\times16$ to $128\times128$ super-resolution. For Inpainting, we use the publicly available implementation\footnote{\url{https://github.com/Janspiry/Palette-Image-to-Image-Diffusion-Models}} of Palette\cite{palette} pretrained on CelebA-HQ to perform hybrid inpainting of images with size $256\times256$. We use center inpainting with an area of $128\times128$. For the sampling method, we generate $200$ variations of each image. To speed up the sampling process we use a cosine scheduler with $200$ denoising steps. For G-Con\textit{ffusion} we use the $128\times128$ ``Guided Diffusion`` model\cite{ADM} pretrained on ImageNet\cite{imagenet}. We copy the original generation head of the DM for predicting $\tilde{u}$ and $\tilde{l}$. We performed a grid search on the optimal noising step, see App.~\ref{app:impl_details}.

\noindent \textbf{Data.} We use the test set of CelebA-HQ\cite{celeba_hq} containing $2000$ samples with a $1000/200/800$ split (finetune/validation/test). The validation is used for computing $\hat{\lambda}$ and performing hyperparameter search, we report the results on the test set using the $\hat{\lambda}$ calculated on the validation set. 

\noindent \textbf{Intervals.} For all of our experiments we use $\alpha=\delta=0.1$, a lower quantile of $0.05$ and an upper quantile of $0.95$.

\subsection{Super-resolution} 

\begin{wrapfigure}{r}{0.5\textwidth}
    \includegraphics[width=0.9\linewidth]{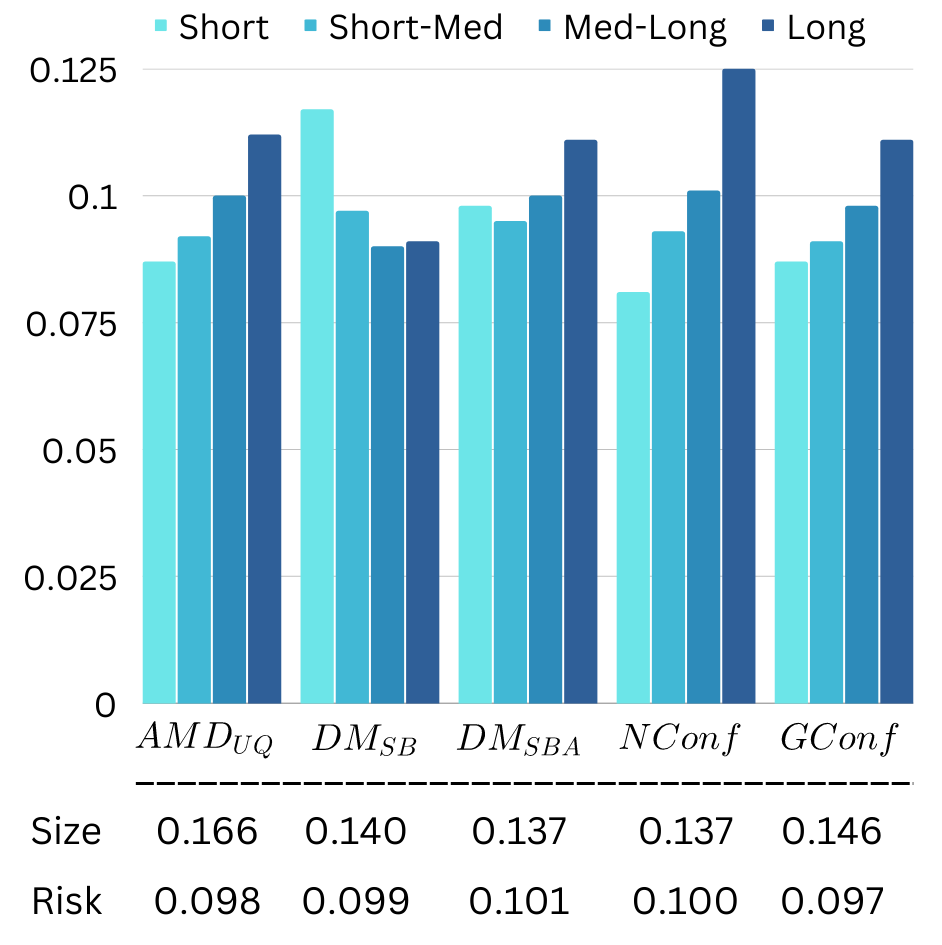}
    \caption{\textit{\textbf{Super-resolution:}} All models satisfy the RCPS definition. All models outperform $ADM_{UQ}$, while $DM_{SBA}$ and N-Con\textit{ffusion} produce slightly tighter intervals than the rest. $DM_{SBA}$ achieves the best stratified coverage }
\label{fig:sr_quantitative}
\end{wrapfigure}

As super-resolution aims to restore fine details, where the low-resolution structure is known, only a small amount of noise is required when diffusing the input image. We can see that all of our methods outperform $ADM_{UQ}$ in terms of interval size while being relatively close to each other; $DM_{SBA}$ and N-Con\textit{ffusion} slightly outperform the rest. The size-stratified metric shows that $DM_{SBA}$ provides the best stratified coverage. G-Con\textit{ffusion} did not pretrain for super-resolution tasks and was not pretrained on the specific data, yet it does not fall far behind the rest of the models, acting as a good alternative in cases where a dedicated model is absent. We report the results in Fig.~\ref{fig:sr_quantitative}. In Fig.~\ref{fig:sr_fig} we can see that N-Con\textit{ffusion} provides meaningful bounds, covering a wide range when needed. Examining the error map, we can see that the errors are mostly in areas with high frequencies; inspecting the interval sizes we see areas the model is uncertain about. 

In contrast to super-resolution, where a signal exists across the input image, the inpainting task contains a large area with no signal. As such, stronger noise is needed when diffusing the input image. Moreover, the constructed intervals are wider than those of the super-resolution ones. Here, N-Con\textit{ffusion} excels in terms of interval size, outperforming the rest by a wide margin. All of our models outperform $ADM_{UQ}$ by a wide margin. All methods perform comparably in terms of stratified risk. We report the results in Fig.~\ref{fig:inpainting_quantitative}. In Fig.~\ref{fig:compare} we visually compare the different methods; $ADM_{UQ}$ produces blurry bounds, although $DM_{SB}$ generates the sharpest intervals, the estimated bounds may contain artifacts. N-Con\textit{ffusion} combines the best of both worlds, constructing the tightest intervals while maintaining realistic-looking bounds. In Fig.~\ref{fig:inpainting_fig} we can see that N-Con\textit{ffusion} provides meaningful bounds, covering a wide range when needed.



\subsection{Is Diffusion Pretraining Necessary?}
G-Con\textit{ffusion} achieved competitive results on both tasks without being trained for related tasks. This raises the question of whether pretraining on a large dataset e.g., ImageNet is sufficient without denoising training? We ablate the roles of the architecture, the pretraining dataset, and the diffusion pretraining. To this end, we finetune an image segmentation architecture FPN\cite{fpn} initialized with a pretrained discriminative encoder.

\begin{figure}[t]
\begin{subfigure}{0.5\textwidth}
\includegraphics[width=0.9\linewidth]{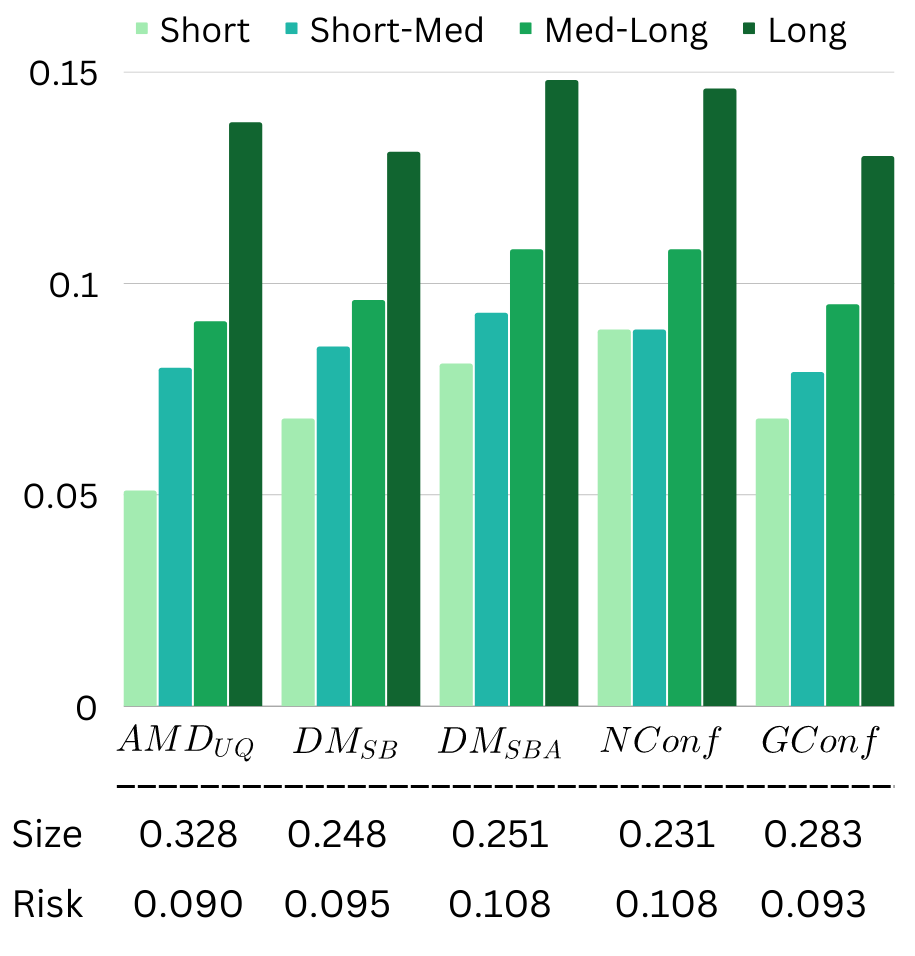}
\caption{\textit{\textbf{Inpainting}}}
\label{fig:inpainting_quantitative}
\end{subfigure}
\begin{subfigure}{0.5\textwidth}
\includegraphics[width=0.9\linewidth]{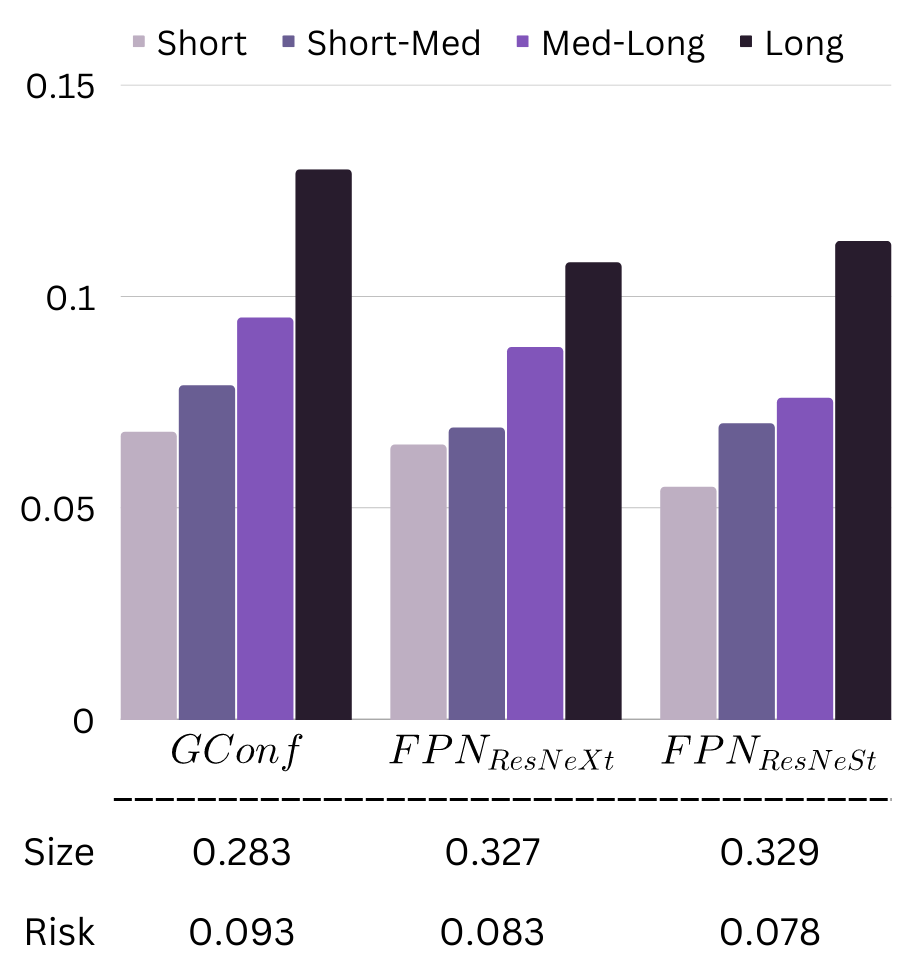}
\caption{\textit{\textbf{Is Diffusion Pretraining Necessary?}}}
\label{fig:ablations_quantitative}
\end{subfigure}
\caption{\textit{\textbf{Inpainting (left):}} All models outperform $ADM_{UQ}$, although N-Con\textit{ffusion} strays slightly from the controlling risk, it excels in terms of interval size.\\
\textit{\textbf{Is Diffusion Pretraining Necessary (right)?}}
  Although all methods satisfy the risk-controlling definition, the intervals they produce are outperformed by diffusion pretraining
}
\end{figure}

The finetuning objective is the same as that of G-Con\textit{ffusion}. We use two different encoders: i) ResNeXt\cite{resnext} comparable in size to ADM ($466M$ parameters for ResNeXt and $422M$ for ADM) that pretrained on the Instagram dataset\cite{instagram_dataset} ii) ResNeSt\cite{resnest} ($108M$ parameters) pretrained on ImageNet. 
Although all methods hold under the risk-controlling definition, the ablated methods produce wider intervals than all our other diffusion-based methods. We report the results for the inpainting task in Fig.~\ref{fig:ablations_quantitative}, for super-resolution ablation see App.~\ref{app:ablation}.

\section{Discussion}
\noindent \textbf{Semantic concept intervals.} This work constructed confidence intervals for low-level vision tasks for which the output is in pixel space. Our work can be used for constructing confidence intervals for more semantic concepts e.g., scene depth or facial age. An added value of constructing confidence intervals for more semantic concepts is that the distribution of solutions often suffers from less variation. The solutions therefore often possess more compact, unimodal distributions, allowing for tighter bounds.

\noindent \textbf{Intervals for unconditional generation.} While we focused in this paper on conditional generation, our work has implications for unconditional generation too. For example, we can use our method to construct confidence intervals for the final generated image (at time $0$), conditioned on the generated image at some intermediate time $t$. This may give us a peak at the statistical properties of the final result - at the cost of only a single forward step. This would be much faster than the full diffusion process. One application can be early stopping of the diffusion process in case the result is undesired.

\noindent \textbf{Adaptive calibration.} During calibration, we seek a single $\hat{\lambda}$ that minimizes the intervals for every pixel of the entire calibration set (while satisfying Def.~\ref{def:rcps}. Such simple parametrization of the calibration process is not strictly necessary. Some more expressive parametrizations are possible e.g., a per-pixel $\lambda$ which would take into account spatially varying aspects of images. This could be necessary when extending our applications to video or 3D scenes. Estimating more parameters may require larger validation sets. 

\noindent \textbf{Number of modes.} Continuous confidence intervals are very effective when the true distribution is unimodal. However, as the distribution of solutions becomes multimodal, their effectiveness is reduced as they are forced to become wide enough to capture the multiple modes. The region between the modes, which may have zero-support is also included in the interval, making it less informative. As multimodal distributions are common, dealing with this setting should be explored in future work.

\section{Conclusion}

This paper tackled confidence bound prediction for image-to-image translation tasks. We first demonstrated that sampling from diffusion models, combined with conformal prediction, outperforms the previously leading method, quantile regression. To accelerate inference speed and interval performance, we presented Con\textit{ffusion}, an approach that combined the best of both worlds. Our method decouples the denoising model from the diffusion process, instead finetuning it using a quantile regression loss. We showed that our method achieves the top performance to date, is fast, and has theoretical guarantees.    


\clearpage
\bibliographystyle{splncs04}
\bibliography{egbib}

\begin{thebibliography}{10}
\providecommand{\url}[1]{\texttt{#1}}
\providecommand{\urlprefix}{URL }
\providecommand{\doi}[1]{https://doi.org/#1}

\bibitem{intro2conformal}
Angelopoulos, A.N., Bates, S.: A gentle introduction to conformal prediction
  and distribution-free uncertainty quantification. arXiv preprint
  arXiv:2107.07511  (2021)

\bibitem{im2imuq}
Angelopoulos, A.N., Kohli, A.P., Bates, S., Jordan, M., Malik, J., Alshaabi,
  T., Upadhyayula, S., Romano, Y.: Image-to-image regression with
  distribution-free uncertainty quantification and applications in imaging. In:
  International Conference on Machine Learning. pp. 717--730. PMLR (2022)

\bibitem{blended_latent_diffusion}
Avrahami, O., Fried, O., Lischinski, D.: Blended latent diffusion. arXiv
  preprint arXiv:2206.02779  (2022)

\bibitem{blended_diffusion}
Avrahami, O., Lischinski, D., Fried, O.: Blended diffusion for text-driven
  editing of natural images. In: Proceedings of the IEEE/CVF Conference on
  Computer Vision and Pattern Recognition. pp. 18208--18218 (2022)

\bibitem{bates2021distribution}
Bates, S., Angelopoulos, A., Lei, L., Malik, J., Jordan, M.: Distribution-free,
  risk-controlling prediction sets. Journal of the ACM (JACM)  \textbf{68}(6),
  1--34 (2021)

\bibitem{come_closer}
Chung, H., Sim, B., Ye, J.C.: Come-closer-diffuse-faster: Accelerating
  conditional diffusion models for inverse problems through stochastic
  contraction. In: Proceedings of the IEEE/CVF Conference on Computer Vision
  and Pattern Recognition. pp. 12413--12422 (2022)

\bibitem{imagenet}
Deng, J., Dong, W., Socher, R., Li, L.J., Li, K., Fei-Fei, L.: Imagenet: A
  large-scale hierarchical image database. In: 2009 IEEE Conference on Computer
  Vision and Pattern Recognition. pp. 248--255 (2009).
  \doi{10.1109/CVPR.2009.5206848}

\bibitem{ADM}
Dhariwal, P., Nichol, A.: Diffusion models beat gans on image synthesis.
  Advances in Neural Information Processing Systems  \textbf{34},  8780--8794
  (2021)

\bibitem{conformal_noise}
Einbinder, B.S., Bates, S., Angelopoulos, A.N., Gendler, A., Romano, Y.:
  Conformal prediction is robust to label noise. arXiv preprint
  arXiv:2209.14295  (2022)

\bibitem{limits_conformal}
Foygel~Barber, R., Candes, E.J., Ramdas, A., Tibshirani, R.J.: The limits of
  distribution-free conditional predictive inference. Information and
  Inference: A Journal of the IMA  \textbf{10}(2),  455--482 (2021)

\bibitem{text_inv}
Gal, R., Alaluf, Y., Atzmon, Y., Patashnik, O., Bermano, A.H., Chechik, G.,
  Cohen-Or, D.: An image is worth one word: Personalizing text-to-image
  generation using textual inversion. arXiv preprint arXiv:2208.01618  (2022)

\bibitem{GAN}
Goodfellow, I., Pouget-Abadie, J., Mirza, M., Xu, B., Warde-Farley, D., Ozair,
  S., Courville, A., Bengio, Y.: Generative adversarial networks.
  Communications of the ACM  \textbf{63}(11),  139--144 (2020)

\bibitem{prompt2prompt}
Hertz, A., Mokady, R., Tenenbaum, J., Aberman, K., Pritch, Y., Cohen-Or, D.:
  Prompt-to-prompt image editing with cross attention control. arXiv preprint
  arXiv:2208.01626  (2022)

\bibitem{DDPM}
Ho, J., Jain, A., Abbeel, P.: Denoising diffusion probabilistic models.
  Advances in Neural Information Processing Systems  \textbf{33},  6840--6851
  (2020)

\bibitem{cascaded}
Ho, J., Saharia, C., Chan, W., Fleet, D.J., Norouzi, M., Salimans, T.: Cascaded
  diffusion models for high fidelity image generation. J. Mach. Learn. Res.
  \textbf{23},  47--1 (2022)

\bibitem{hoeffding}
Hoeffding, W.: Probability inequalities for sums of bounded random variables.
  Journal of the American Statistical Association  \textbf{58}(301),  13--30
  (1963), \url{http://www.jstor.org/stable/2282952}

\bibitem{celeba_hq}
Karras, T., Aila, T., Laine, S., Lehtinen, J.: Progressive growing of gans for
  improved quality, stability, and variation. arXiv preprint arXiv:1710.10196
  (2017)

\bibitem{EDM}
Karras, T., Aittala, M., Aila, T., Laine, S.: Elucidating the design space of
  diffusion-based generative models. arXiv preprint arXiv:2206.00364  (2022)

\bibitem{ffhq}
Karras, T., Laine, S., Aila, T.: A style-based generator architecture for
  generative adversarial networks. In: Proceedings of the IEEE/CVF conference
  on computer vision and pattern recognition. pp. 4401--4410 (2019)

\bibitem{quantile_regression}
Koenker, R., Bassett~Jr, G.: Regression quantiles. Econometrica: journal of the
  Econometric Society pp. 33--50 (1978)

\bibitem{srdiff}
Li, H., Yang, Y., Chang, M., Chen, S., Feng, H., Xu, Z., Li, Q., Chen, Y.:
  Srdiff: Single image super-resolution with diffusion probabilistic models.
  Neurocomputing  \textbf{479},  47--59 (2022)

\bibitem{fpn}
Lin, T.Y., Doll{\'a}r, P., Girshick, R., He, K., Hariharan, B., Belongie, S.:
  Feature pyramid networks for object detection. In: Proceedings of the IEEE
  conference on computer vision and pattern recognition. pp. 2117--2125 (2017)

\bibitem{repaint}
Lugmayr, A., Danelljan, M., Romero, A., Yu, F., Timofte, R., Van~Gool, L.:
  Repaint: Inpainting using denoising diffusion probabilistic models. In:
  Proceedings of the IEEE/CVF Conference on Computer Vision and Pattern
  Recognition. pp. 11461--11471 (2022)

\bibitem{instagram_dataset}
Mahajan, D., Girshick, R., Ramanathan, V., He, K., Paluri, M., Li, Y.,
  Bharambe, A., Van Der~Maaten, L.: Exploring the limits of weakly supervised
  pretraining. In: Proceedings of the European conference on computer vision
  (ECCV). pp. 181--196 (2018)

\bibitem{diffae}
Preechakul, K., Chatthee, N., Wizadwongsa, S., Suwajanakorn, S.: Diffusion
  autoencoders: Toward a meaningful and decodable representation. In: IEEE
  Conference on Computer Vision and Pattern Recognition (CVPR) (2022)

\bibitem{dalle2}
Ramesh, A., Dhariwal, P., Nichol, A., Chu, C., Chen, M.: Hierarchical
  text-conditional image generation with clip latents. arXiv preprint
  arXiv:2204.06125  (2022)

\bibitem{ldm}
Rombach, R., Blattmann, A., Lorenz, D., Esser, P., Ommer, B.: High-resolution
  image synthesis with latent diffusion models. In: Proceedings of the IEEE/CVF
  Conference on Computer Vision and Pattern Recognition. pp. 10684--10695
  (2022)

\bibitem{unet}
Ronneberger, O., Fischer, P., Brox, T.: U-net: Convolutional networks for
  biomedical image segmentation. In: International Conference on Medical image
  computing and computer-assisted intervention. pp. 234--241. Springer (2015)

\bibitem{dreambooth}
Ruiz, N., Li, Y., Jampani, V., Pritch, Y., Rubinstein, M., Aberman, K.:
  Dreambooth: Fine tuning text-to-image diffusion models for subject-driven
  generation  (2022)

\bibitem{palette}
Saharia, C., Chan, W., Chang, H., Lee, C., Ho, J., Salimans, T., Fleet, D.,
  Norouzi, M.: Palette: Image-to-image diffusion models. In: ACM SIGGRAPH 2022
  Conference Proceedings. pp. 1--10 (2022)

\bibitem{imagen}
Saharia, C., Chan, W., Saxena, S., Li, L., Whang, J., Denton, E., Ghasemipour,
  S.K.S., Ayan, B.K., Mahdavi, S.S., Lopes, R.G., et~al.: Photorealistic
  text-to-image diffusion models with deep language understanding. arXiv
  preprint arXiv:2205.11487  (2022)

\bibitem{sr3}
Saharia, C., Ho, J., Chan, W., Salimans, T., Fleet, D.J., Norouzi, M.: Image
  super-resolution via iterative refinement. IEEE Transactions on Pattern
  Analysis and Machine Intelligence  (2022)

\bibitem{disentabgled_semantic_intervals}
Sankaranarayanan, S., Angelopoulos, A.N., Bates, S., Romano, Y., Isola, P.:
  Semantic uncertainty intervals for disentangled latent spaces. arXiv preprint
  arXiv:2207.10074  (2022)

\bibitem{score_matching}
Sohl-Dickstein, J., Weiss, E., Maheswaranathan, N., Ganguli, S.: Deep
  unsupervised learning using nonequilibrium thermodynamics. In: International
  Conference on Machine Learning. pp. 2256--2265. PMLR (2015)

\bibitem{DDIM}
Song, J., Meng, C., Ermon, S.: Denoising diffusion implicit models. arXiv
  preprint arXiv:2010.02502  (2020)

\bibitem{conformal_feat}
Teng, J., Wen, C., Zhang, D., Bengio, Y., Gao, Y., Yuan, Y.: Predictive
  inference with feature conformal prediction. arXiv preprint arXiv:2210.00173
  (2022)

\bibitem{resnext}
Xie, S., Girshick, R., Doll{\'a}r, P., Tu, Z., He, K.: Aggregated residual
  transformations for deep neural networks. In: Proceedings of the IEEE
  conference on computer vision and pattern recognition. pp. 1492--1500 (2017)

\bibitem{resnest}
Zhang, H., Wu, C., Zhang, Z., Zhu, Y., Lin, H., Zhang, Z., Sun, Y., He, T.,
  Mueller, J., Manmatha, R., et~al.: Resnest: Split-attention networks. In:
  Proceedings of the IEEE/CVF Conference on Computer Vision and Pattern
  Recognition. pp. 2736--2746 (2022)

\end{thebibliography}


\appendix

\clearpage

\section{Calibration}
\label{app:calibration}
Following \cite{im2imuq}, we perform the calibration by forming a Hoeffding's upper-confidence bound,

\begin{equation}
    \widehat{R}^+(\lambda)= \frac{1}{n}\sum\limits_{i=1}^nL(\mathcal{T}_{\lambda}(x),y)+\sqrt{\frac{1}{2n}\log\frac{1}{\delta}}.
\end{equation}
In \cite{hoeffding} it is shown that this is a valid bound that holds for $\mathbb{P}\left[\widehat{R}^+(\lambda) < R(\lambda) \right] < \delta$. Meaning, we can use $\widehat{R}^+(\lambda)$ to choose the smallest $\lambda$ that satisfies the definition of RCPS. For the proof and further discussion regarding these properties, see \cite{bates2021distribution}. The pseudocode for the above algorithm can be found at \cite{im2imuq}.

\section{Implementation Details}
\label{app:impl_details}
\noindent \textbf{Noise Level.} We used grid search to choose which amount of noise to use (i.e. the step to diffuse the input image to). For the super-resolution tasks, for SR3 we use $t=0$, for guided-diffusion, we use $t=3$. For Inpainting tasks, for palette we use $t=155$, for guided-diffusion we use $t=20$.

\noindent \textbf{Noise Injection Method.} The noise injection method (i.e. input image diffusion) varies based on the model used. For SR3 and Palette, the noise is concatenated to the input image. For guided-diffusion, the noise is added to the input image (point-wise summation).

\begin{wrapfigure}{r}{0.5\textwidth}
    \includegraphics[width=1.0\linewidth]{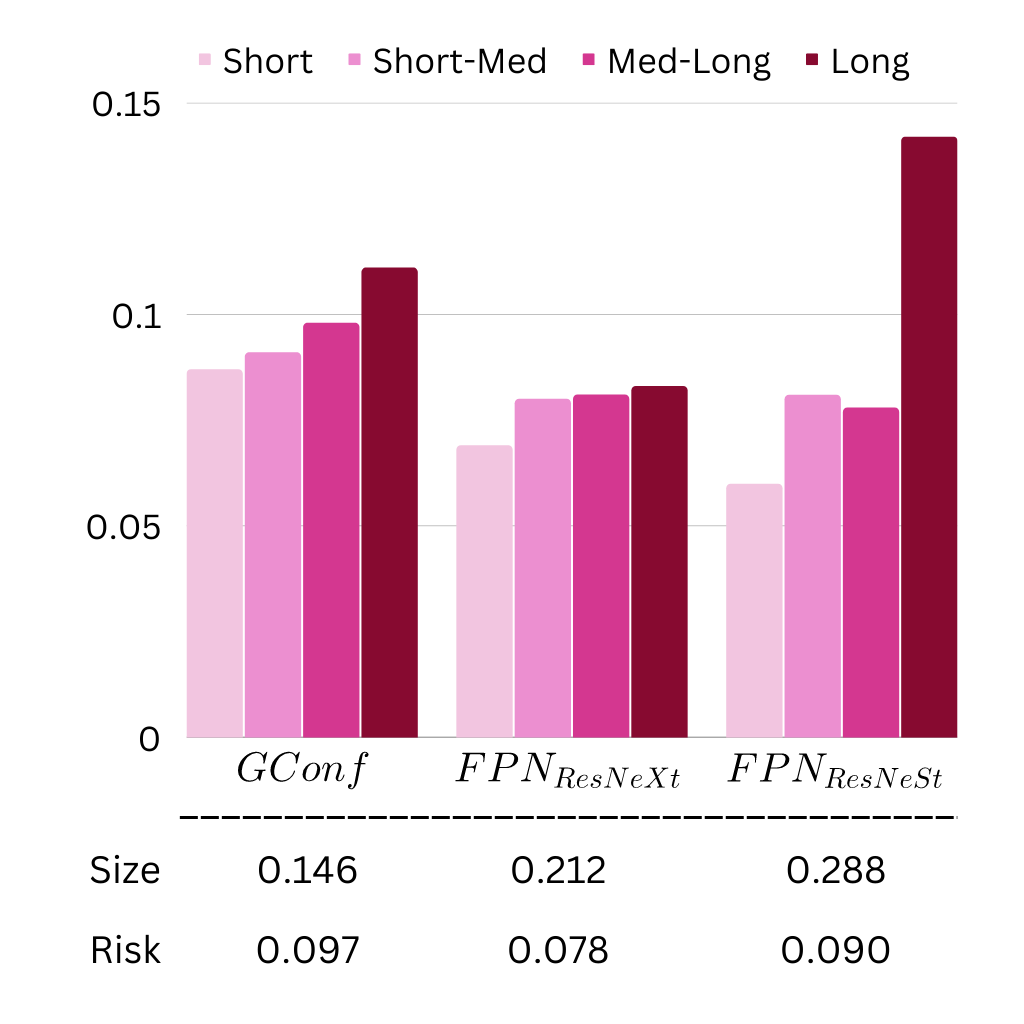}
  \caption{\textit{\textbf{Is Diffusion Pretraining Necessary?}}
  Although all methods satisfy the risk-controlling definition, the intervals they produce are outperformed by diffusion pretraining}
\label{fig:ablations_quantitative_sr}
\end{wrapfigure}

\noindent \textbf{Ablation.} The "Is diffusion pretraining necessary" ablation was performed using an FPN with two different backbones. For all of the experiments we used the publicly available\footnote{\url{https://github.com/qubvel/segmentation_models.pytorch}} implementation of ``Pytorch Segmentation Models``. We used the pretrained backbones ``resnext101\_32x32d`` and ``timm-resnest269e``.

\section{Is Diffusion Pretraining Necessary?}
\label{app:ablation}

Following the ablation shown in the paper, here we show the results for the super-resolution task. As seen in Fig.~\ref{fig:ablations_quantitative_sr}, although all methods hold under the risk-controlling definition, the ablated methods produce wider intervals than all our other diffusion-based methods.


\section{Additional Visual Comparisons}
\label{app:comp}
\begin{figure*}[t!]
\begin{tabular}{c@{\hskip1pt}c@{\hskip1pt}c@{\hskip1pt}c@{\hskip1pt}c@{\hskip1pt}c@{\hskip1pt}c@{\hskip1pt}}
\rotatebox[ origin=c]{90}{$ADM_{UQ}$} &
 \includegraphics[align=c, width=0.16\linewidth]{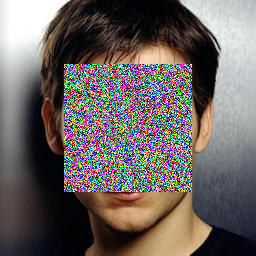} & 
\includegraphics[align=c, width=0.16\linewidth]{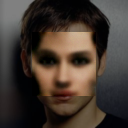} & 
\includegraphics[align=c, width=0.16\linewidth]{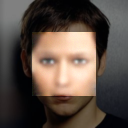} & 
\includegraphics[align=c, width=0.16\linewidth]{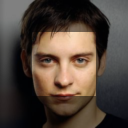} & 
\includegraphics[align=c, width=0.16\linewidth]{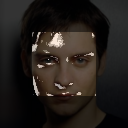} &
\includegraphics[align=c, width=0.16\linewidth]{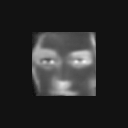}\\\addlinespace

\rotatebox[ origin=c]{90}{$DM_{SB}$} &
 \includegraphics[align=c, width=0.16\linewidth]{figs/compare1/SB_509_masked_smpl.png} & 
\includegraphics[align=c, width=0.16\linewidth]{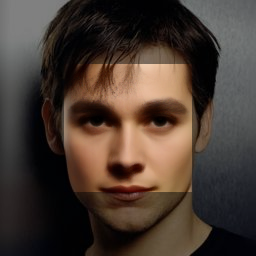} & 
\includegraphics[align=c, width=0.16\linewidth]{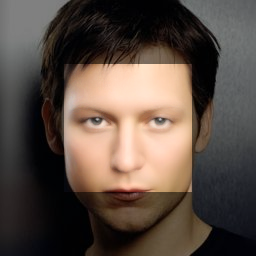} & 
\includegraphics[align=c, width=0.16\linewidth]{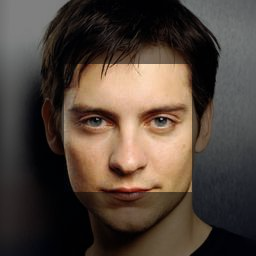} & 
\includegraphics[align=c, width=0.16\linewidth]{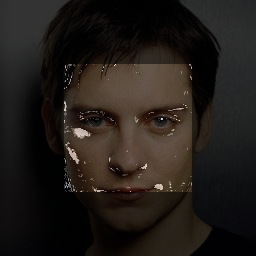} &
\includegraphics[align=c, width=0.16\linewidth]{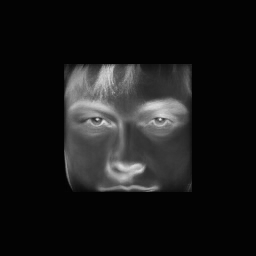}\\\addlinespace

\rotatebox[ origin=c]{90}{$DM_{SBA}$} &
\includegraphics[align=c, width=0.16\linewidth]{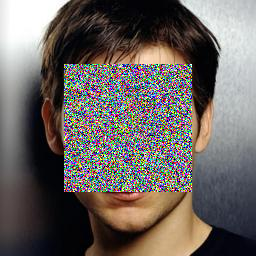} & 
\includegraphics[align=c, width=0.16\linewidth]{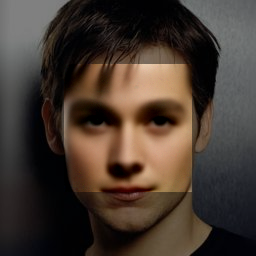} & 
\includegraphics[align=c, width=0.16\linewidth]{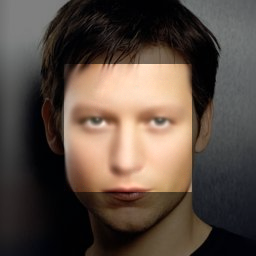} & 
\includegraphics[align=c, width=0.16\linewidth]{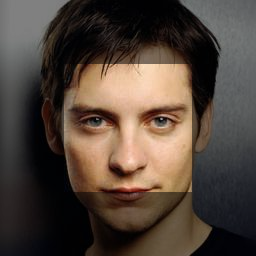} & 
\includegraphics[align=c, width=0.16\linewidth]{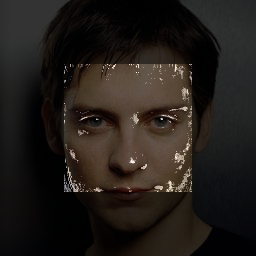}&
\includegraphics[align=c, width=0.16\linewidth]{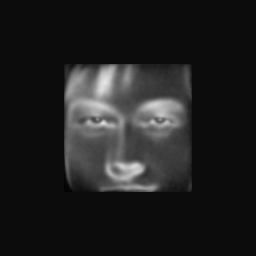}\\\addlinespace

\rotatebox[ origin=c]{90}{N-Con\textit{ffusion}} &
\includegraphics[align=c, width=0.16\linewidth]{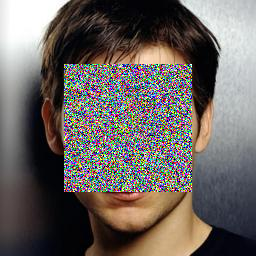} & 
\includegraphics[align=c, width=0.16\linewidth]{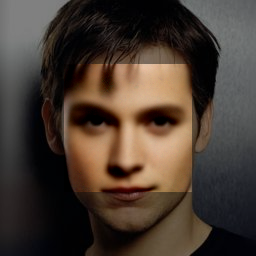} & 
\includegraphics[align=c, width=0.16\linewidth]{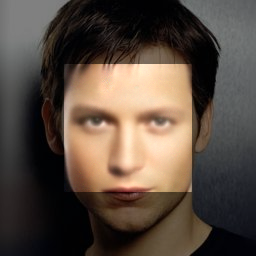} & 
\includegraphics[align=c, width=0.16\linewidth]{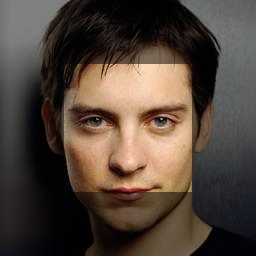} & 
\includegraphics[align=c, width=0.16\linewidth]{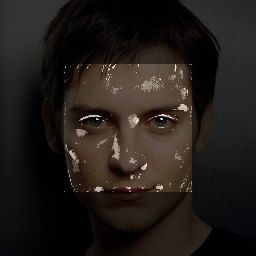} &
\includegraphics[align=c, width=0.16\linewidth]{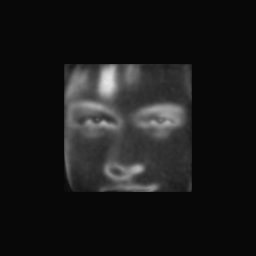}\\\addlinespace

\rotatebox[ origin=c]{90}{G-Con\textit{ffusion}} &
\includegraphics[align=c, width=0.16\linewidth]{figs/compare1/SB_509_masked_smpl.png} & 
\includegraphics[align=c, width=0.16\linewidth]{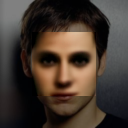} & 
\includegraphics[align=c, width=0.16\linewidth]{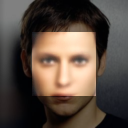} & 
\includegraphics[align=c, width=0.16\linewidth]{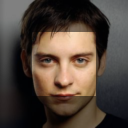} & 
\includegraphics[align=c, width=0.16\linewidth]{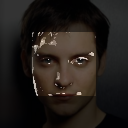} &
\includegraphics[align=c, width=0.16\linewidth]{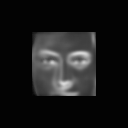}\\\addlinespace
 & Input & Lower Bound & Upper Bound & Ground Truth & Error & Interval Size \\
\end{tabular}
 \caption{\textit{\textbf{Comparing the Different Methods:}} We compare the different methods  on the inpainting task. $ADM_{UQ}$ produces blurry bounds, also apparent from the smoother interval heatmap. Although $DM_{SB}$ generates the sharpest intervals, the estimated bounds may contain artifacts (e.g. hair strands). N-Con\textit{ffusion} combines the best of both worlds, generating sharp intervals while maintaining realistic bounds}
\label{fig:compare1}
\end{figure*}

\begin{figure*}[t!]
\begin{tabular}{c@{\hskip1pt}c@{\hskip1pt}c@{\hskip1pt}c@{\hskip1pt}c@{\hskip1pt}c@{\hskip1pt}c@{\hskip1pt}}

\rotatebox[ origin=c]{90}{$ADM_{UQ}$} &
 \includegraphics[align=c,width=0.16\linewidth]{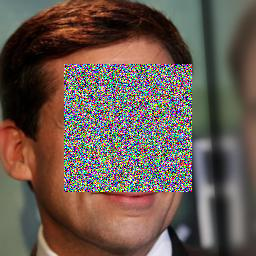} & 
\includegraphics[align=c,width=0.16\linewidth]{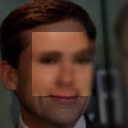} & 
\includegraphics[align=c,width=0.16\linewidth]{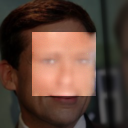} & 
\includegraphics[align=c,width=0.16\linewidth]{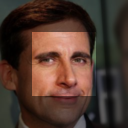} & 
\includegraphics[align=c,width=0.16\linewidth]{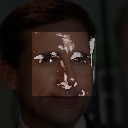} &
\includegraphics[align=c,width=0.16\linewidth]{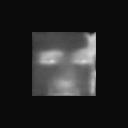}\\\addlinespace

\rotatebox[ origin=c]{90}{$DM_{SB}$} &
 \includegraphics[align=c,width=0.16\linewidth]{figs/compare2/SB_519_masked_smpl.png} & 
\includegraphics[align=c,width=0.16\linewidth]{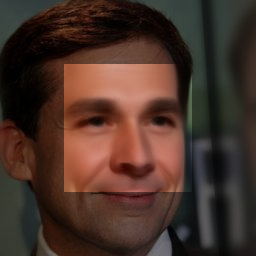} & 
\includegraphics[align=c,width=0.16\linewidth]{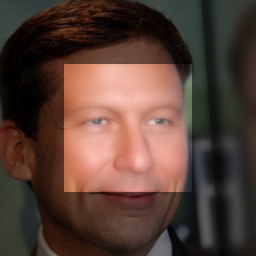} & 
\includegraphics[align=c,width=0.16\linewidth]{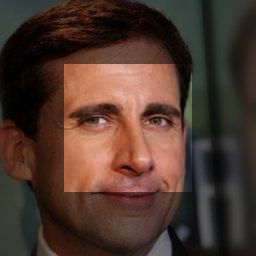} & 
\includegraphics[align=c,width=0.16\linewidth]{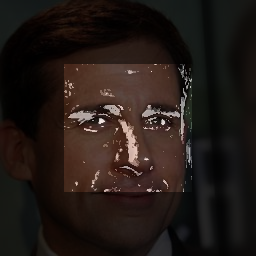} &
\includegraphics[align=c,width=0.16\linewidth]{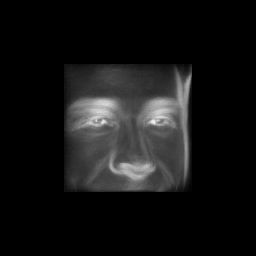}\\\addlinespace

\rotatebox[ origin=c]{90}{$DM_{SBA}$} &
\includegraphics[align=c, width=0.16\linewidth]{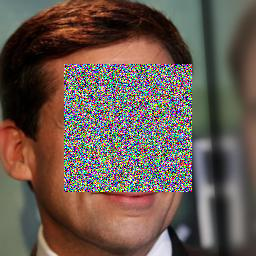} & 
\includegraphics[align=c, width=0.16\linewidth]{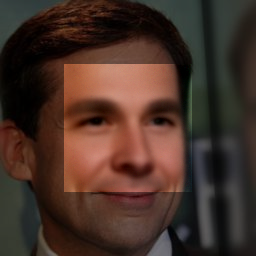} & 
\includegraphics[align=c, width=0.16\linewidth]{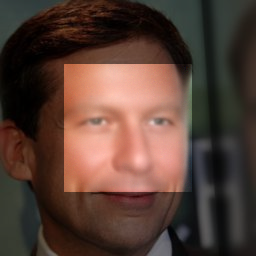} & 
\includegraphics[align=c, width=0.16\linewidth]{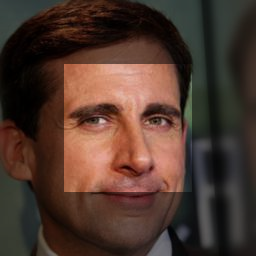} & 
\includegraphics[align=c, width=0.16\linewidth]{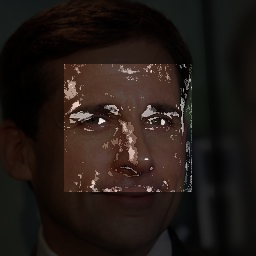}&
\includegraphics[align=c, width=0.16\linewidth]{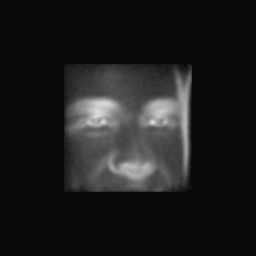}\\\addlinespace

\rotatebox[ origin=c]{90}{N-Con\textit{ffusion}} &
\includegraphics[align=c, width=0.16\linewidth]{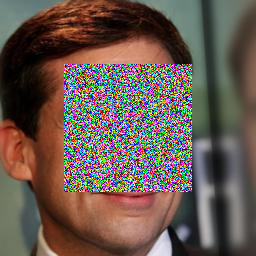} & 
\includegraphics[align=c, width=0.16\linewidth]{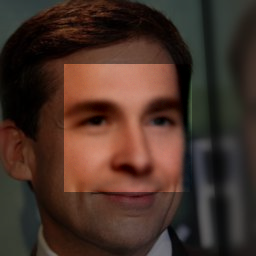} & 
\includegraphics[align=c, width=0.16\linewidth]{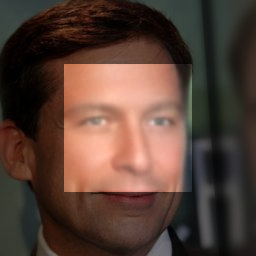} & 
\includegraphics[align=c, width=0.16\linewidth]{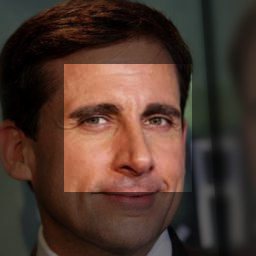} & 
\includegraphics[align=c, width=0.16\linewidth]{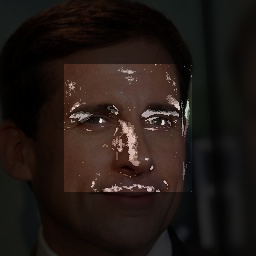} &
\includegraphics[align=c, width=0.16\linewidth]{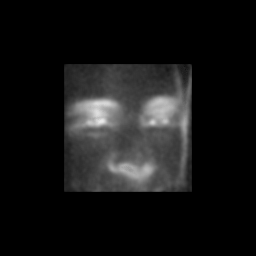}\\\addlinespace

\rotatebox[ origin=c]{90}{G-Con\textit{ffusion}} &
\includegraphics[align=c, width=0.16\linewidth]{figs/compare2/SB_519_masked_smpl.png} & 
\includegraphics[align=c, width=0.16\linewidth]{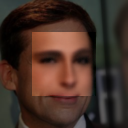} & 
\includegraphics[align=c, width=0.16\linewidth]{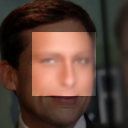} & 
\includegraphics[align=c, width=0.16\linewidth]{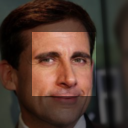} & 
\includegraphics[align=c, width=0.16\linewidth]{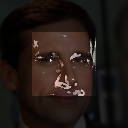} &
\includegraphics[align=c, width=0.16\linewidth]{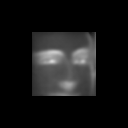}\\\addlinespace
 & Input & Lower Bound & Upper Bound & Ground Truth & Error & Interval Size \\
\end{tabular}
 \caption{\textit{\textbf{Comparing the Different Methods:}} We compare the different methods on the inpainting task. $ADM_{UQ}$ produces blurry bounds, also apparent from the smoother interval heatmap. $DM_{SBA}$ closely approximates $DM_{SB}$}
\label{fig:compare2}
\end{figure*}

\begin{figure*}[t!]

\begin{tabular}{c@{\hskip1pt}c@{\hskip1pt}c@{\hskip1pt}c@{\hskip1pt}c@{\hskip1pt}c@{\hskip1pt}c@{\hskip1pt}}
\rotatebox[ origin=c]{90}{$ADM_{UQ}$} &
 \includegraphics[align=c, width=0.16\linewidth]{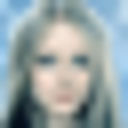} & 
\includegraphics[align=c, width=0.16\linewidth]{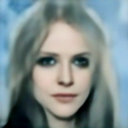} & 
\includegraphics[align=c, width=0.16\linewidth]{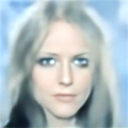} & 
\includegraphics[align=c, width=0.16\linewidth]{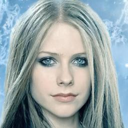} & 
\includegraphics[align=c, width=0.16\linewidth]{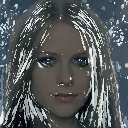} &
\includegraphics[align=c, width=0.16\linewidth]{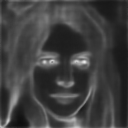}\\\addlinespace

\rotatebox[ origin=c]{90}{$DM_{SB}$} &
\includegraphics[align=c, width=0.16\linewidth]{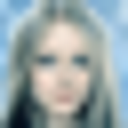} & 
\includegraphics[align=c, width=0.16\linewidth]{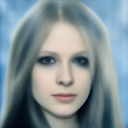} & 
\includegraphics[align=c, width=0.16\linewidth]{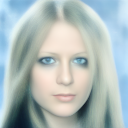} & 
\includegraphics[align=c, width=0.16\linewidth]{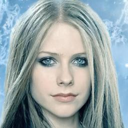} & 
\includegraphics[align=c, width=0.16\linewidth]{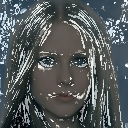} &
\includegraphics[align=c, width=0.16\linewidth]{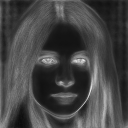}\\\addlinespace

\rotatebox[ origin=c]{90}{$DM_{SBA}$} &
\includegraphics[align=c, width=0.16\linewidth]{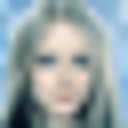} & 
\includegraphics[align=c, width=0.16\linewidth]{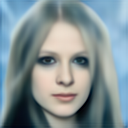} & 
\includegraphics[align=c, width=0.16\linewidth]{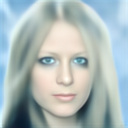} & 
\includegraphics[align=c, width=0.16\linewidth]{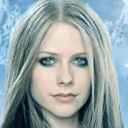} & 
\includegraphics[align=c, width=0.16\linewidth]{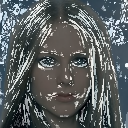} &
\includegraphics[align=c, width=0.16\linewidth]{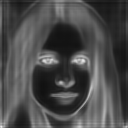}\\\addlinespace

\rotatebox[ origin=c]{90}{N-Con\textit{ffusion}} &
\includegraphics[align=c, width=0.16\linewidth]{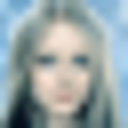} & 
\includegraphics[align=c, width=0.16\linewidth]{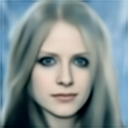} & 
\includegraphics[align=c, width=0.16\linewidth]{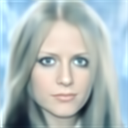} & 
\includegraphics[align=c, width=0.16\linewidth]{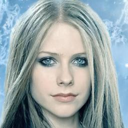} & 
\includegraphics[align=c, width=0.16\linewidth]{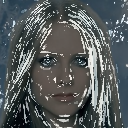} &
\includegraphics[align=c, width=0.16\linewidth]{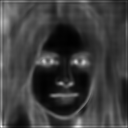}\\\addlinespace

\rotatebox[ origin=c]{90}{G-Con\textit{ffusion}} &
\includegraphics[align=c, width=0.16\linewidth]{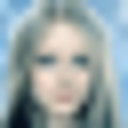} & 
\includegraphics[align=c, width=0.16\linewidth]{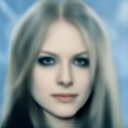} & 
\includegraphics[align=c, width=0.16\linewidth]{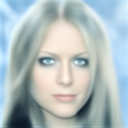} & 
\includegraphics[align=c, width=0.16\linewidth]{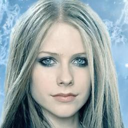} & 
\includegraphics[align=c, width=0.16\linewidth]{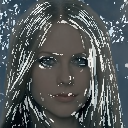} &
\includegraphics[align=c, width=0.16\linewidth]{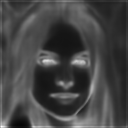}\\\addlinespace
 & Input & Lower Bound & Upper Bound & Ground Truth & Error & Interval Size \\
\end{tabular}
 \caption{\textit{\textbf{Comparing the Different Methods:}} We compare the different methods on the super-resolution task. $ADM_{UQ}$ produces blurry bounds, also apparent from the smoother interval heatmap. Although $DM_{SB}$ generates the sharpest intervals, the estimated bounds may contain artifacts (e.g. right eye). N-Con\textit{ffusion} combines the best of both worlds, generating sharp intervals while maintaining realistic bounds. The error occurs mostly in high-frequency areas}
\label{fig:compare_sr1}
\end{figure*}

\begin{figure*}[t!]
\begin{tabular}{c@{\hskip1pt}c@{\hskip1pt}c@{\hskip1pt}c@{\hskip1pt}c@{\hskip1pt}c@{\hskip1pt}c@{\hskip1pt}}
\rotatebox[ origin=c]{90}{$ADM_{UQ}$} &
 \includegraphics[align=c, width=0.16\linewidth]{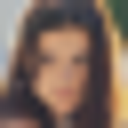} & 
\includegraphics[align=c, width=0.16\linewidth]{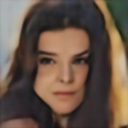} & 
\includegraphics[align=c, width=0.16\linewidth]{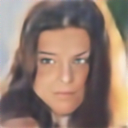} & 
\includegraphics[align=c, width=0.16\linewidth]{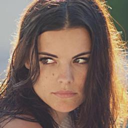} & 
\includegraphics[align=c, width=0.16\linewidth]{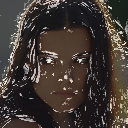} &
\includegraphics[align=c, width=0.16\linewidth]{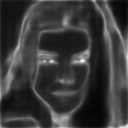}\\\addlinespace

\rotatebox[ origin=c]{90}{$DM_{SB}$} &
\includegraphics[align=c, width=0.16\linewidth]{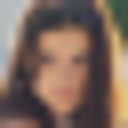} & 
\includegraphics[align=c, width=0.16\linewidth]{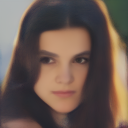} & 
\includegraphics[align=c, width=0.16\linewidth]{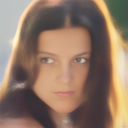} & 
\includegraphics[align=c, width=0.16\linewidth]{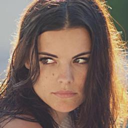} & 
\includegraphics[align=c, width=0.16\linewidth]{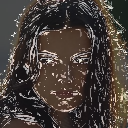} &
\includegraphics[align=c, width=0.16\linewidth]{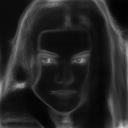}\\\addlinespace

\rotatebox[ origin=c]{90}{$DM_{SBA}$} &
\includegraphics[align=c, width=0.16\linewidth]{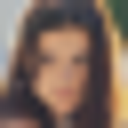} & 
\includegraphics[align=c, width=0.16\linewidth]{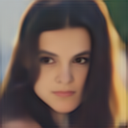} & 
\includegraphics[align=c, width=0.16\linewidth]{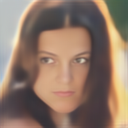} & 
\includegraphics[align=c, width=0.16\linewidth]{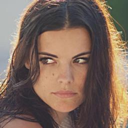} & 
\includegraphics[align=c, width=0.16\linewidth]{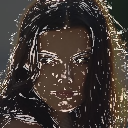} &
\includegraphics[align=c, width=0.16\linewidth]{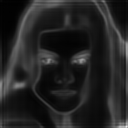}\\\addlinespace

\rotatebox[ origin=c]{90}{N-Con\textit{ffusion}} &
\includegraphics[align=c, width=0.16\linewidth]{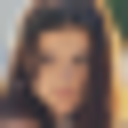} & 
\includegraphics[align=c, width=0.16\linewidth]{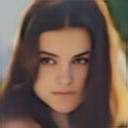} & 
\includegraphics[align=c, width=0.16\linewidth]{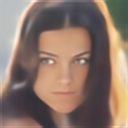} & 
\includegraphics[align=c, width=0.16\linewidth]{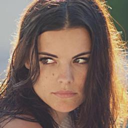} & 
\includegraphics[align=c, width=0.16\linewidth]{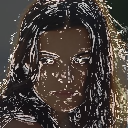} &
\includegraphics[align=c, width=0.16\linewidth]{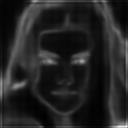}\\\addlinespace

\rotatebox[ origin=c]{90}{G-Con\textit{ffusion}} &
\includegraphics[align=c, width=0.16\linewidth]{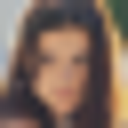} & 
\includegraphics[align=c, width=0.16\linewidth]{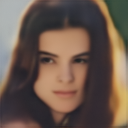} & 
\includegraphics[align=c, width=0.16\linewidth]{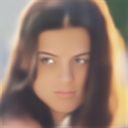} & 
\includegraphics[align=c, width=0.16\linewidth]{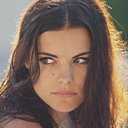} & 
\includegraphics[align=c, width=0.16\linewidth]{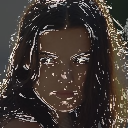} &
\includegraphics[align=c, width=0.16\linewidth]{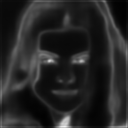}\\\addlinespace
 & Input & Lower Bound & Upper Bound & Ground Truth & Error & Interval Size \\
\end{tabular}
 \caption{\textit{\textbf{Comparing the Different Methods:}} We compare the different methods on the super-resolution task. The $ADM_{UQ}$ bounds fail to capture the original semantics , also apparent from the smoother interval heatmap. Although $DM_{SB}$ generates the sharpest intervals, it sometimes fails to capture the full range of possibilities (e.g. eye color). N-Con\textit{ffusion} combines the best of both worlds, generating sharp intervals while maintaining realistic bounds (e.g. larger eye color range). The error occurs mostly in high-frequency areas}
\label{fig:compare_sr2}
\end{figure*}

In Fig.~\ref{fig:compare1} and Fig.~\ref{fig:compare2} we visualize more bound comparisons for the inpainting task. In Fig.~\ref{fig:compare_sr1} and Fig.~\ref{fig:compare_sr2} we visualize bound comparisons for the super-resolution task.

\end{document}